\documentclass[lettersize,journal]{IEEEtran}
\usepackage{amsmath,amsfonts}
\usepackage{algorithmic}
\usepackage{algorithm}
\usepackage{array}
\usepackage[caption=false,font=normalsize,labelfont=sf,textfont=sf]{subfig}
\usepackage{textcomp}
\usepackage{stfloats}
\usepackage{url}
\usepackage{verbatim}
\usepackage{graphicx}
\usepackage{cite}
\usepackage{hyperref}
\hypersetup{
    colorlinks=true,
    linkcolor=blue,
    filecolor=blue,
    urlcolor=blue,
    citecolor=blue,
}

\usepackage{xcolor}
\usepackage{comment}

\begin{document}

\title{Analysis and experiments of the dissipative Twistcar: \\ direction reversal and asymptotic approximations}

\author{Rom~Levy,
        Ari~Dantus,
        Zitao~Yu,
        and~Yizhar~Or
\thanks{This work was supported by Israel Science
Foundation under grant no. 1382/23.}
\thanks{R. Levy, A. Dantus and Y. Or are with the Department of Mechanical Engineering, Technion - Israel Institute of Technology, Haifa 3200003, Israel.}
\thanks{Z. Yu is with the Department of Aerospace and Mechanical Engineering, University of Southern California, USA. The work of Z. Yu on this research has been conducted as part of his MSc Thesis at Technion.}
\thanks{\textit{Corresponding author: Y. Or} (e-mail: \href{mailto:izi@technion.ac.il}{izi@technion.ac.il})}
}



\maketitle

\begin{abstract}
Underactuated wheeled vehicles are commonly studied as nonholonomic systems with periodic actuation. Twistcar is a classical example inspired by a riding toy, which has been analyzed using a planar model of a dynamical system with nonholonomic constraints. Most of the previous analyses did not account for energy dissipation due to frictional resistance. In this work, we study a theoretical two-link model of the Twistcar while incorporating dissipation due to rolling resistance. We obtain asymptotic expressions for the system's small-amplitude steady-state periodic dynamics, which reveals the possibility of reversing the direction of motion upon varying the geometric and mass properties of the vehicle. Next, we design and construct a robotic prototype of the Twistcar whose center-of-mass position can be shifted by adding and removing a massive block, enabling experimental demonstration of the Twistcar’s direction reversal phenomenon. We also conduct parameter fitting for the frictional resistance in order to improve agreement with experiments.
\end{abstract}

\begin{IEEEkeywords}
Robotics, Under-actuated systems, nonholonomic dynamics, asymptotic analysis.
\end{IEEEkeywords}

\section{Introduction}
\IEEEPARstart{U}{nderactuated} locomotive systems are systems whose number of degrees-of-freedom (DoF) is greater than the number of its controlled inputs. Such systems often incorporate nonholonomic constraints \cite{bloch_basic_2003, ne_mark_dynamics_2004}, which restrict permissible velocity directions without decreasing the effective number of the system's DoF. Examples of such systems can be found in terrestrial motion, such as the Snakeboard \cite{ostrowski_nonholonomic_1994} and the three-wheeled snake \cite{raz2026tro}, or for swimming in fluid, including Purcell's swimmer \cite{purcell_life_1977, wiezel_optimization_2016} and inertia-dominated swimmers \cite{kanso_locomotion_2005, virozub_planar_2019, hatton_geometric_2013}.

Based on the number of constraints and DoF, those systems can be classified as purely kinematic \cite{buchanan_geometric_2021} or mixed kinematic-dynamic systems \cite{ostrowski_mechanics_1995}. Purely kinematic systems, such as the three-linked kinematic snake \cite{shammas_geometric_2007} and Purcell's swimmer, contain an equal number of nonholonomic constraints and passive (unactuated) DoF of body motion. This property leads to body motion that is governed only by the geometric changes of the robot's actuated inputs of shape variables. When the number of constraints is smaller, changes in the system's momentum influence its movement alongside the kinematic constraints \cite{shammas_towards_2007}. A classical example of a mixed system is Chaplygin’s sleigh \cite{stanchenko_non-holonomic_1989}, and its extensions with oscillatory actuation, such as the Chaplygin's beanie \cite{david_kelly_proportional_2012} and Chaplygin sleigh with a moving mass \cite{osborne_steering_2005}. Other examples consist of wheeled rolling board toys such as the Snakeboard and Roller-racer \cite{krishnaprasad_oscillations_2001, yang_geometric_2025}, which is also known in the literature as the Landfish \cite{dear_mechanics_2014}, Land-shark \cite{bazzi_motion_2017} or Twistcar \cite{chakon_analysis_2017} - all are renamed variants of the same toy, originally called Roller-racer.

\begin{figure*}[h]
\centering
\includegraphics[width=\textwidth,clip]{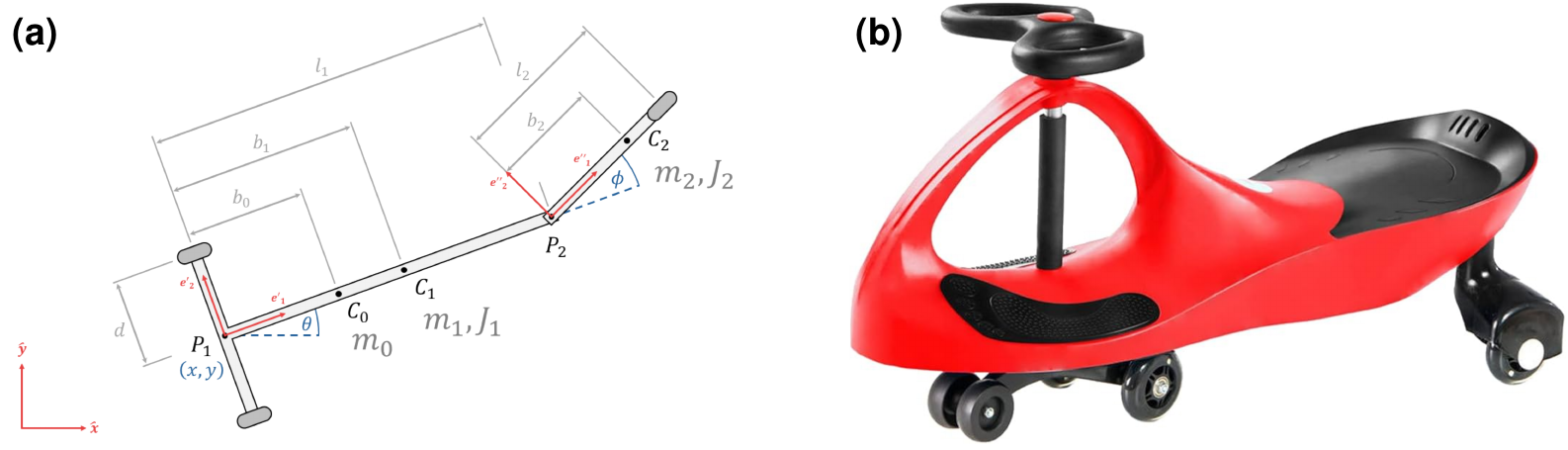}
\caption{The Twistcar vehicle. (a) Simplified two-link model. (b) A Twistcar toy example.} \label{fig1}
\end{figure*}

The Twistcar, shown in Figure~\ref{fig1}, is an underactuated two-linked vehicle with passive wheels. The vehicle is actuated under time-periodic input of the relative angle between the two links. Previous work included a perturbation analysis \cite{nayfeh_perturbation_2008} on a point-mass model \cite{chakon_analysis_2017} and was extended by considering a more general model, incorporating mass and inertia of the two links \cite{halvani_nonholonomic_2022}. It was demonstrated theoretically that a direction reversal in the vehicle's movement direction is possible by changing the geometric properties of the model. This phenomenon was analyzed in \cite{chakon_analysis_2017, halvani_nonholonomic_2022, bazzi_motion_2017} using different approaches. Along with other similar models \cite{bizyaev_exotic_2018}, it has been shown that the non-dissipative nature of the nonoholonomic constraints leads to a diverging amplitude of the vehicle's oscillatory velocity and orientation angle. Importantly, this unbounded behavior does not hold in experiments, as reported in \cite{halvani_nonholonomic_2022, kilin_theoretical_2025}.

\IEEEpubidadjcol
This theoretically-unbounded behavior arises from the absence of energy dissipation in the system, since the forces maintaining the nonholonomic no-skid constraints do not perform mechanical work. Several methods have been considered to incorporate the wheels' skid, a behavior also shown in the experiments, into the mathematical model: Some works considered replacing the nonholonomic constraints with a Coulomb's friction model \cite{fedonyuk2017stick, yona_wheeled_2019, halvani_nonholonomic_2022}. Although physically plausible, this hybrid model is numerically sensitive and cannot be studied using analytical tools such as perturbation expansion \cite{chakon_analysis_2017}, due to its non-smoothness. The work \cite{bazzi_motion_2017}, considered a skid-angle model \cite{sidek_dynamic_2008, bazzi_novel_2014} which mimics skid velocities without adding dissipation to the system. While this model results in smooth dynamics, it still leads to unbounded diverging oscillations due to a lack of dissipation. Another approach introduces dissipation to the system while keeping the nonholonomic no-skid constraints. Recent work \cite{kilin_theoretical_2025, yang_geometric_2025} suggests using viscous dissipation in rolling resistance of the wheels, and another work used a similar approach on a Snakeboard model \cite{dear_snakeboard_2015}. This approach leads to bounded behavior with smooth dynamics.

In this work, we revisit the Twistcar dynamics using the latter assumption of viscous dissipation due to wheels' rolling resistance, while preserving the nonholonomic no-skid constraints at the lateral directions. { Inspired by practical actuation of related toys, and motivated by the line of research in recent works on nonholonomic underactuated systems, we consider the system's open-loop dynamics under the simplest form of time-periodic input, namely, a sinusoidal signal.} We conduct a small-amplitude analysis of the model using perturbation expansion in order to obtain approximate explicit expressions which are compared to numerical simulations, with excellent agreement for small actuation amplitudes. In addition, we design and construct a robotic prototype of the Twistcar with an open-loop controller of the steering joint's angle. The robot has a modular structure which enables changing the robot's geometry and mass distribution. The perturbation analysis, simulations, and experiments, all result in a bounded oscillatory motion. Using the asymptotic solution, we obtain explicit conditions for direction reversal due to adding or removing a massive block onto the robot's body. The theoretical predictions are corroborated by the motion of our modular robot, thus demonstrating direction reversal in actual motion experiments.

The paper is organized as follows: In section~\ref{sec2} we formulate the dynamic equations of the problem, considering the nonholonomic no-skid constraints and the dissipative rolling resistance of the wheels. Then, in section~\ref{sec3} we conduct numerical simulations demonstrating the bounded behavior of the model and the direction reversal phenomenon. In section~\ref{sec4} we obtain approximate asymptotic expressions for the solution of the vehicle’s forward speed and orientation angle, assuming small-amplitude actuation. In section~\ref{sec5} we present the Twistcar robot and experimental results obtained using it. We estimate the viscous dissipation coefficient of the wheels by using parameter fitting between the experimental and numerical results, and demonstrate direction reversal upon shifting the vehicle’s center of mass, as predicted by the theoretical analysis. In section~\ref{sec6} we discuss additional effects that influence the behavior of the experimental robot - the wheels' skidding and the non-perfectly horizontal terrain on which the experiment was conducted. We show the significance of each effect numerically. Finally, section~\ref{sec7} concludes the work and suggests directions for future extensions.  

\section{Problem formulation} \label{sec2}
We now present formulation of the Twistcar’s dynamic equations of motion, a planar model analogous to that described by \cite{halvani_nonholonomic_2022} as depicted in Figure~\ref{fig1}. The vehicle has two rigid links, each characterized by a total length \(l_i\), a center of mass (CoM) located at a distance \(b_i\) from point \(P_i\), a mass \(m_i\), and a moment of inertia \(J_i\) about its CoM. Two wheels are positioned symmetrically at a lateral distance \(d\) from point \(P_1\), while a third wheel is mounted at the apex of link 2. We introduce a point-mass \(m_0\), located at a distance \(b_0\) from point \(P_1\). We choose the following generalized coordinates:
\begin{equation} \label{Generalized Coordinates}
    \textbf{q} = \begin{bmatrix}x & y & \theta & \phi\end{bmatrix}^T
\end{equation}
Where \( (x, y) \) denote the location of point \(P_1\), \(\theta\) is the orientation angle of link 1 about the global \(\hat{\textbf{x}}\) axis, and \(\phi\) is the relative angle between the links which we refer to as the steering angle. We use the steering angle as the system's input: 
\begin{equation} \label{Input Angle}
    \phi(t) = \phi_0 + \varepsilon \cos{\left( \omega t \right)}
\end{equation}

To enforce the no-skid constraints, we express the wheels' skid velocities in terms of the generalized coordinates:
\begin{equation} \label{Skid Velocity - with Jacobian}
    \textbf{v}_{\perp, i} = \mathbb{J}_{\perp, i}(\textbf{q}) \dot{\textbf{q}}, \quad i=1,2
\end{equation}

Using the Jacobian notation, we define the nonholonomic no-slip constraints in \eqref{Nonholonomic Constraints}, by setting the slip velocities to zero:
\begin{equation} \label{Nonholonomic Constraints}
    \textbf{W}(\textbf{q}) \dot{\textbf{q}} = \textbf{0}
\end{equation}
where the nonholonomic constraints matrix is:
\begin{equation}
    \textbf{W} = 
    \begin{bmatrix}
        - \sin{\theta} & \cos{\theta} & 0 & 0 \\
        - \sin{\left( \theta + \phi \right)} & \cos{\left( \theta + \phi \right)} & l_2 + l_1 \cos{\phi} & l_2
    \end{bmatrix}
\end{equation}

In order to add viscous rolling resistance to the model, we use Rayleigh's dissipation function, as follows. For each wheel, we define the velocity along its permissible roll direction using the Jacobian:
\begin{equation} \label{Roll Velocity - with Jacobian}
    \textbf{v}_{\parallel, i} = \mathbb{J}_i(\textbf{q}) \dot{\textbf{q}}, \quad i=1,2,3
\end{equation}

Using the Jacobian notation, we define Rayleigh's dissipation function as in \eqref{Rayleigh Dissipation Function}, where \(c\) is the dissipation coefficient.
\begin{equation} \label{Rayleigh Dissipation Function}
    \mathcal{R}(\textbf{q}, \dot{\textbf{q}}) = \frac{c}{2} \sum_{i=1}^{3}{ {\textbf{v}_{\parallel, i}}^{T} \textbf{v}_{\parallel, i}} = \frac{c}{2} {\dot{\textbf{q}}}^T \left( \sum_{i=1}^{3}{{\mathbb{J}_i(\textbf{q})}^T \mathbb{J}_i(\textbf{q})} \right) \dot{\textbf{q}}
\end{equation}

We differentiate \eqref{Rayleigh Dissipation Function} with respect to the generalized velocity \(\dot{\textbf{q}}\) to obtain the dissipation contribution to the dynamic equations. Using the Jacobian notation, one obtains:
\begin{equation} \label{Dissipation Vector}
    \textbf{D}(\textbf{q}, \dot{\textbf{q}}) = \frac{\partial \mathcal{R}}{\partial \dot{\textbf{q}}} = c \left( \sum_{i=1}^{3}{{\mathbb{J}_i(\textbf{q})}^T \mathbb{J}_i(\textbf{q})} \right) \dot{\textbf{q}}
\end{equation}

To derive the constrained dynamic equations of motion, we use the Euler–Lagrange method \cite{murray_mathematical_2017}, and include the nonholonomic forces acting on the model. We define the total kinetic energy of the problem \(T\left(\textbf{q},\dot{\textbf{q}}\right)\) and use a planar model with constant potential energy to obtain \eqref{Euler-Lagrange Constrained Equation}.
\begin{equation} \label{Euler-Lagrange Constrained Equation}
    \frac{d}{dt} \left( \frac{\partial T}{\partial \dot{\textbf{q}}} \right) - \frac{\partial T}{\partial \textbf{q}} + \frac{\partial \mathcal{R}}{\partial \dot{\textbf{q}}} = \textbf{F}_q + \textbf{W}^T(\textbf{q}) \boldsymbol{\Lambda}
\end{equation}

where \(\boldsymbol{\Lambda}\) is the vector of forces enforcing the constraints, and \(\textbf{F}_q\) is the vector of generalized forces, which include the actuation torque \(\tau\) at the steering joint.
\begin{equation*}
    \boldsymbol{\Lambda} = \begin{bmatrix} \lambda_1 & \lambda_2 \end{bmatrix}^T, \quad \textbf{F}_q = \begin{bmatrix}0 & 0 & 0 & \tau\end{bmatrix}^T
\end{equation*}

We rearrange \eqref{Euler-Lagrange Constrained Equation} in matrix form and differentiate \eqref{Nonholonomic Constraints} with respect to time. This results in the following differential algebraic system.
\begin{align}
    \textbf{M}(\textbf{q}) \ddot{\textbf{q}} + \textbf{B}(\textbf{q}, \dot{\textbf{q}}) + \textbf{D}(\textbf{q}, \dot{\textbf{q}}) &= \textbf{F}_q + \textbf{W}^T(\textbf{q}) \boldsymbol{\Lambda} \nonumber \\
    \textbf{W}(\textbf{q}) \ddot{\textbf{q}} + \dot{\textbf{W}}(\textbf{q}) \dot{\textbf{q}} &= 0 \label{Dynamic Equations - Matrix Form}
\end{align}
where:
\begin{equation*}
    \textbf{M}(\textbf{q}) = \frac{\partial^2 T}{\partial \dot{\textbf{q}}^2}, \quad \textbf{B}(\textbf{q}, \dot{\textbf{q}}) = \frac{\partial^2 T}{\partial \dot{\textbf{q}} \partial\textbf{q}} \dot{\textbf{q}} - \frac{\partial T}{\partial \textbf{q}}
\end{equation*}
{ The Twistcar's dynamic is thus formulated as a nonholonomic system with a single kinematic input $\phi(t)$, which is essentially similar to the roller-racer \cite{krishnaprasad_oscillations_2001}. As proven in the \underline{ \href{https://users.ics.forth.gr/~tsakiris/Papers/CDCSS_TR_98_4.pdf}{extended preprint version}} of \cite{krishnaprasad_oscillations_2001}, this system is not differnetially flat, and thus it is not amenable to simplified methods of path planning using flat output and dynamic feedback \cite{bloch_basic_2003}.}

\section{Simulations} \label{sec3}
In this section, we reformulate the dynamic equations in order to conduct numerical analysis. It allows us to obtain the system's solution in time for different model parameters, showing the bounded behavior of the model and the direction reversal phenomenon.

\begin{figure*}[!b]
\centering
\includegraphics[width=\textwidth, clip]{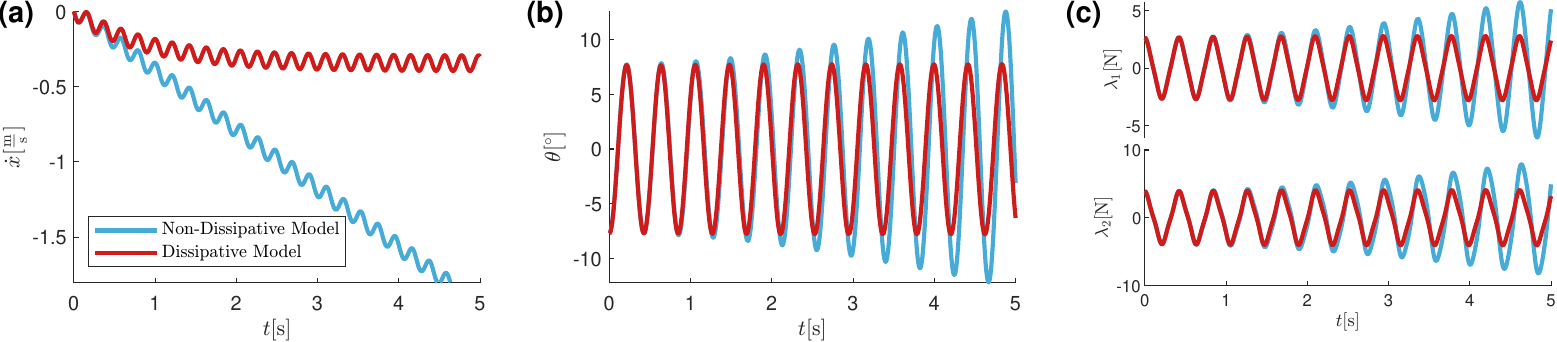} 
\caption{Numeric simulation results for the dissipative and non-dissipative models, using the parameters in \eqref{Input Angle for simulations}, \eqref{Simulation Parameters}. (a) The vehicle's speed \(\dot{x}(t)\) in the world frame. (b) The angle of Link 1 relative to the average direction of progression. (c) Nonholonomic no-skid constraint forces.} \label{fig2}
\end{figure*}

To do so, we divide the generalized coordinates into active coordinates, which we directly control, and passive coordinates.
\begin{equation}
    \textbf{q}_p = \begin{bmatrix}x & y & \theta\end{bmatrix}^T, \quad \textbf{q}_a = \phi
\end{equation}

We decompose the matrices in \eqref{Dynamic Equations - Matrix Form} with respect to the active and passive coordinates:
\begin{align}
    \begin{bmatrix}
        \textbf{M}_{pp} & \textbf{M}_{pa} \\
        {\textbf{M}_{pa}}^T & \textbf{M}_{aa}
    \end{bmatrix} \begin{bmatrix} \ddot{\textbf{q}}_p \\ \ddot{\textbf{q}}_a\end{bmatrix} + \begin{bmatrix} \textbf{B}_p \\ \textbf{B}_a \end{bmatrix} + \begin{bmatrix} \textbf{D}_p \\ \textbf{D}_a \end{bmatrix} &= \begin{bmatrix} \textbf{0}^{3 \times 1} \\ \tau \end{bmatrix} + \begin{bmatrix}{\textbf{W}_p}^T \\ {\textbf{W}_a}^T\end{bmatrix} \boldsymbol{\Lambda} \nonumber \\
    \textbf{W}_p \ddot{\textbf{q}}_p + \textbf{W}_a \ddot{\textbf{q}}_a + \dot{\textbf{W}}_p \dot{\textbf{q}}_p &+ \dot{\textbf{W}}_a \dot{\textbf{q}}_a = 0 \label{Dynamic Equations - Passive-Active}
\end{align}
where:
{\fontsize{8}{9}
\begin{align*}
    \textbf{M}_{pp} \in \mathbb{R}^{3 \times 3}, \quad \textbf{M}_{pa} \in \mathbb{R}^{3 \times 1}, \quad \textbf{M}_{aa} \in \mathbb{R}, \quad \textbf{B}_p \in \mathbb{R}^{3 \times 1},& \quad \textbf{B}_a \in \mathbb{R} \\
    \textbf{D}_p \in \mathbb{R}^{3 \times 1}, \quad \textbf{D}_a \in \mathbb{R}, \quad \textbf{W}_p \in \mathbb{R}^{2 \times 3}, \quad \textbf{W}_a \in \mathbb{R}^{2 \times 1},& \quad \tau \in \mathbb{R}
\end{align*}
}
We rearrange \eqref{Dynamic Equations - Passive-Active} in a new matrix form, where the unknown vector consists of the passive coordinates' acceleration, the actuation torque, and the constraint forces.
\begin{equation} \label{Dynamic Equations - Simulations}
    {\fontsize{8}{9}
    \begin{bmatrix}
        \ddot{\textbf{q}}_p \\ \tau \\ \boldsymbol{\Lambda}
    \end{bmatrix} = - \begin{bmatrix}
        \textbf{M}_{pp} & \textbf{0}^{3 \times 1} & - {\textbf{W}_p}^T \\
        {\textbf{M}_{pa}}^T & -1 & - {\textbf{W}_a}^T \\
        \textbf{W}_p & \textbf{0}^{2 \times 1} & \textbf{0}^{2 \times 2}
    \end{bmatrix}^{-1}\begin{bmatrix}
        \textbf{M}_{pa} \ddot{\textbf{q}}_a + \textbf{B}_p + \textbf{D}_p \\
        \textbf{M}_{aa} \ddot{\textbf{q}}_a + \textbf{B}_a + \textbf{D}_a\\
        \textbf{W}_a \ddot{\textbf{q}}_a + \dot{\textbf{W}}_p \dot{\textbf{q}}_p + \dot{\textbf{W}}_a \dot{\textbf{q}}_a
    \end{bmatrix}
    }
\end{equation}

Using \eqref{Dynamic Equations - Simulations}, we numerically solve the first three equations to find the time solutions of the passive coordinates \(\textbf{q}_p(t)\). All simulations were conducted using MATLAB's \textit{ode45} function. To obtain the input torque \(\tau(t)\) and the constraint forces \(\boldsymbol{\Lambda}(t)\), we substitute the solution \(\textbf{q}_p(t)\) into the last three equations in \eqref{Dynamic Equations - Simulations}.

To determine the movement direction and velocity of the vehicle, we examine its velocity along the average direction of progression in the world frame. We first simulate under zero initial conditions and obtain the mean value of the orientation angle in steady-state \(\overline{\theta}\). By subtracting \(\overline{\theta}\) from \(\theta(t)\), we obtain a solution such that it oscillates around zero, and the vehicle's net displacement is along the world frame's \(\hat{x}\) axis.

Simulations were conducted using \eqref{Dynamic Equations - Simulations}, with an input angle as in \eqref{Input Angle}, for:
\begin{equation} \label{Input Angle for simulations}
    \phi_0 = 0, \quad \varepsilon = \frac{\pi}{6}, \quad \omega = 15 \mathrm{\left[ rad \, / \, s\right]}
\end{equation}

The parameters of the problems were taken under the assumption that each link is a slender rod with symmetric mass distribution.
\begin{align}
    m_0 = 0, \quad &m_1 = 1 \mathrm{[Kg]}, \quad m_2 = 0.3 \mathrm{[Kg]}, \quad c = 0.5 \mathrm{[N \cdot s / m]}, \nonumber \\
    l_1 = &0.3 \mathrm{[m]}, \quad l_2 = 0.1 \mathrm{[m]}, \quad d = 0.05 \mathrm{[m]} \label{Simulation Parameters} \\
    &\quad b_i = l_i / 2, \quad J_i = m_i {l_i}^2 / 12 \qquad \text{for } i = 1,2 \nonumber
\end{align}

In Figure~\ref{fig2} we show numerical time solutions for both dissipative and non-dissipative models. For the non-dissipative model, we use the same formulation and parameters, except for taking \(c=0\). In contrast to the unphysical unbounded behavior of the non-dissipative model, the dissipative model has a bounded oscillatory steady-state behavior.

\begin{figure}[h]
\centering
\includegraphics[width= 0.45\textwidth, clip]{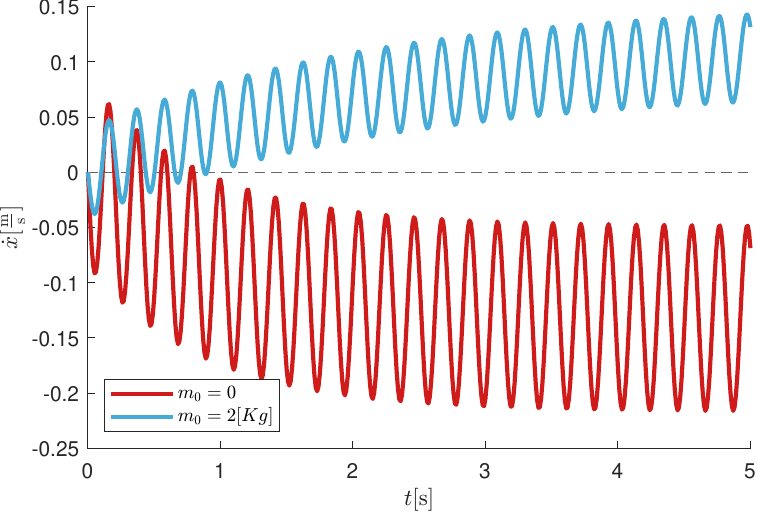}
\caption{Direction reversal demonstration using numeric simulation with the parameters in \eqref{Input Angle}, \eqref{Simulation Parameters}, achieved by adding a point-mass to the vehicle.} \label{fig3}
\end{figure}

Next, we demonstrate direction reversal by changing the geometry and mass distribution of the model. We modify the parameters presented in \eqref{Simulation Parameters} by changing the length of link 2 to \(l_2 = \mathrm{0.2[m]}\) and adding a point-mass \(m_0\) at \(b_0 = 0.05 \mathrm{[m]}\). 

In Figure~\ref{fig3} we show how changing the mass distribution of the vehicle influences its movement direction. Without the point-mass \(m_0=0\), the vehicle moves backward \(v(t)<0\), toward link 1. By adding a point-mass \(m_0\), the vehicle's speed changes from negative to positive, meaning the vehicle moves forward toward link 2. Using this result, we design an experimental setup that captures the direction reversal phenomenon, as shown below in Section~\ref{sec5.4}.

\section{Asymptotic analysis} \label{sec4}
In this section, we conduct an approximated asymptotic analysis of the model's dynamics. To conduct the analysis, we first reduce the number of variables by enforcing the constraints. Then, we obtain an asymptotic approximation using perturbation analysis \cite{nayfeh_perturbation_2008}. 

We define a body-frame velocity vector:
\begin{equation}
    \textbf{v}_b = \begin{bmatrix}v_{\parallel} & v_{\perp} & \dot{\theta} & \dot{\phi}\end{bmatrix}^T
\end{equation}
Where \(v_{\parallel}\) and \(v_{\perp}\) are the velocity components of point \(P_1\) along \(\textbf{e}_1'\) and \(\textbf{e}_2'\) directions, i.e. parallel and perpendicular to the vehicle's body.

In \eqref{Generalized Velocities Transformation} we present the transformation from the body-frame velocities to the generalized velocities of the problem.
\begin{equation} \label{Generalized Velocities Transformation}
    \dot{\textbf{q}} = \textbf{R}_b \textbf{v}_b, \quad \textbf{R}_b = \begin{bmatrix}
        \cos{\theta} & - \sin{\theta} & 0 & 0 \\
        \sin{\theta} &  \cos{\theta} & 0 & 0 \\
        0 & 0 & 1 & 0 \\
        0 & 0 & 0 & 1
    \end{bmatrix}
\end{equation}

Next, we use the nonholonomic no-skid constraints in \eqref{Nonholonomic Constraints} to reduce the dimensionality of the vehicle's dynamics. In \eqref{Nonholonomic Constraints - Body Velocities} we rewrite the constraints in terms of the body-frame velocities.
\begin{align} \label{Nonholonomic Constraints - Body Velocities}
    \textbf{W} \dot{\textbf{q}} &= \left( \textbf{W} \textbf{R}_b \right) \left( {\textbf{R}_b}^{-1} \dot{\textbf{q}} \right) = \textbf{W}_b \textbf{v}_b \\
    &=\begin{bmatrix} 0 && 1 && 0 && 0 \\ -\sin{\phi} && \cos{\phi} && l_1 \cos{\phi} + l_2 && l_2 \end{bmatrix} \begin{bmatrix} v_{\parallel} \\ v_{\perp} \\ \dot{\theta} \\ \dot{\phi} \end{bmatrix} = \textbf{0} \nonumber
\end{align}

This defines two relations between the body-frame velocities. The first directly dictates that the lateral velocity \(v_{\perp}\) is zero. We use the second row of \eqref{Nonholonomic Constraints - Body Velocities} in order to eliminate \(\dot{\theta}\), and express the equations of motion using a reduced vector of unconstrained velocities, defined as:
\begin{equation}
    \textbf{v}_r = \begin{bmatrix}v_{\parallel} & \dot{\phi}\end{bmatrix}^T
\end{equation}

We rewrite \eqref{Nonholonomic Constraints - Body Velocities} as a relation between the reduced velocities and the body-frame velocities in \eqref{Body Velocities Reduction}.
\begin{equation} \label{Body Velocities Reduction}
    \textbf{v}_b = \textbf{S}_r \textbf{v}_r, \quad \textbf{S}_r =
    \begin{bmatrix}
        1 & 0 \\
        0 & 0 \\
        \frac{\sin{\phi}}{l_2 + l_1 \cos{\phi}} & - \frac{l_2}{l_2 + l_1 \cos{\phi}} \\
        0 & 1
    \end{bmatrix}
\end{equation}

We substitute \eqref{Body Velocities Reduction} into \eqref{Generalized Velocities Transformation} and obtain a relation between the generalized velocity and the reduced velocity vectors:
\begin{equation} \label{Full Velocity Transformation}
    \dot{\textbf{q}} = \textbf{S} { \textbf{v}_r}, \quad \textbf{S} = \textbf{R}_b \textbf{S}_r
\end{equation}

Applying this relation in \eqref{Dynamic Equations - Matrix Form} leads to \eqref{Dynamic Equations - Reduced} below. We emphasize that this reduction naturally enforces the nonholonomic no-skid constraints in \eqref{Nonholonomic Constraints - Body Velocities}, nullifying all terms involving \(\textbf{W}\).
\begin{equation} \label{Dynamic Equations - Reduced}
    \textbf{M}_r(\phi) \dot{\textbf{v}}_r + \textbf{B}_r(\phi, \textbf{v}_r) + \textbf{D}_r(\phi, \textbf{v}_r) = \textbf{F}_r
\end{equation}
where:
\begin{align*}
    \textbf{M}_r(\phi) = \textbf{S}^T \textbf{M} \textbf{S} \in \mathbb{R}^{2 \times 2},& \quad \textbf{F}_r = \begin{bmatrix}0 & \tau\end{bmatrix}^T, \\
    \textbf{B}_r(\phi, \textbf{v}_r) = \textbf{S}^T \textbf{M} \dot{\textbf{S}} \textbf{v}_b& + \textbf{S}^T \textbf{B} \in \mathbb{R}^{2 \times 1}
\end{align*}

The first equation in \eqref{Dynamic Equations - Reduced} is decoupled from the second one, and has the structure:
\begin{equation} \label{Asymptotic Analysis - Equation Form}
    \dot{v} = f(v, \phi, \dot{\phi}, \ddot{\phi})
\end{equation}
Note that from \eqref{Asymptotic Analysis - Equation Form} on, we refer to the forward speed of the vehicle \(v_{\parallel}\) as \(v\). We use \eqref{Asymptotic Analysis - Equation Form} to solve for \(v(t)\) under prescribed input \(\phi(t)\).

To reduce the number of parameters in the problem, we merge the point-mass with link 1. We denote the previous parameters of link 1 with \(^*\), so that their modified values are:
\begin{equation} \label{Incorporate Point-Mass into Link 1}
    m_1 = m_1^* + m_0, \quad J_1 = J_1^* + m_0 \left(b_0 - b_1\right)^2
\end{equation}

Now we nondimensionalize the problem by defining the nondimensional parameters:
\begin{equation} \label{Nondimensional Parameters}
    \alpha = \frac{l_2}{l_1}, \sigma = \frac{d}{l_1}, \kappa = \frac{m_2}{m_1}, \beta_i = \frac{b_i}{l_i}, \eta_i = \frac{J_i}{m_i {l_i}^2} \quad i = 1,2
\end{equation}

We define the characteristic time of the problem \(t_c = m_1 / c\), and denote \(\tilde{t} = t / t_c\) and \(\tilde{\omega} = \omega t_c\). The nondimensional time-dependent variables are:
\begin{equation} \label{Nondimensional Time-Dependent Variables}
    \tilde{v} = v \frac{t_c}{l_1}, \quad \tilde{v}' = \frac{d \tilde{v}}{d \tilde{t}} = \dot{v} \frac{{t_c}^2}{l_1}, \quad \phi' = \dot{\phi} t_c, \quad \phi'' = \ddot{\phi} {t_c}^2
\end{equation}

From now on, we analyze the nondimensional problem and drop the tildes for brevity.

\subsection{Symmetric input}
We first consider the problem under a symmetric time-periodic input of the steering angle, meaning that the input angle defined in \eqref{Input Angle} has a mean value \(\phi_0 = 0\). We write \eqref{Asymptotic Analysis - Equation Form} in a nondimensional form:
\begin{equation} \label{Asymptotic Analysis - Nondimensional Equation Form}
    \resizebox{!}{.85em}{ \(v'(t) = \tilde{f}\left(v(t), \varepsilon \cos{(\omega t)}, - \varepsilon \omega \sin{(\omega t)}, - \varepsilon \omega^2 \cos{(\omega t)}\right) \)}
\end{equation}

We use perturbation analysis assuming small angle amplitude \(\varepsilon \ll 1\), and expand \(v(t)\) as a power series in \(\varepsilon\):
\begin{equation} \label{Symmetric - Velocity Expansion}
    v(t) = v_0(t) + \varepsilon v_1(t) + \varepsilon^2 v_2(t) + \ldots
\end{equation}

After substituting \eqref{Symmetric - Velocity Expansion} into \eqref{Asymptotic Analysis - Nondimensional Equation Form}, We expand \(\tilde{f}\) as a power series in \(\varepsilon\) and arrange the expression in powers of \(\varepsilon\). Under zero initial conditions, the leading and first-order solutions are nullified \(v_0(t) = v_1(t) \equiv 0\), and \(v(t)\) is an even-power series. The second-order ODE is in the form of:
\begin{equation} \label{Symmetric - Velocity ODE}
    v_2'(t) = a_1 + a_2 \, v_2(t) + a_3 \sin{(2 \omega t)} + a_4 \cos{(2 \omega t)}
\end{equation}
where \(a_1, a_2, a_3\) appear in Appendix \hyperref[tab2]{A}. We solve \eqref{Symmetric - Velocity ODE} and obtain:
\begin{equation} \label{Symmetric - Velocity Solution}
    v_2(t) = b_1 + b_2 \mathrm{e}^{-\frac{3t}{1 + \kappa}} + b_3 \sin{\left(2 \omega t \right)} + b_4 \cos{\left(2 \omega t \right)}
\end{equation}
where:
\begin{align*}
    b_1 = \alpha  \omega ^2 (\alpha  \kappa  \left(\left(\beta _2-2\right) \beta _2+\eta _2\right)+&\kappa  \left(\alpha +\beta_2-1\right) \\
     +\alpha  \beta _1&-\beta _1^2-\eta _1) / \left(6(\alpha + 1)^2\right)
\end{align*}
and \(b_2,b_3,b_4\) appear in Appendix \hyperref[tab2]{A}.

\begin{figure*}[!t]
\centering
\includegraphics[width=\textwidth, clip]{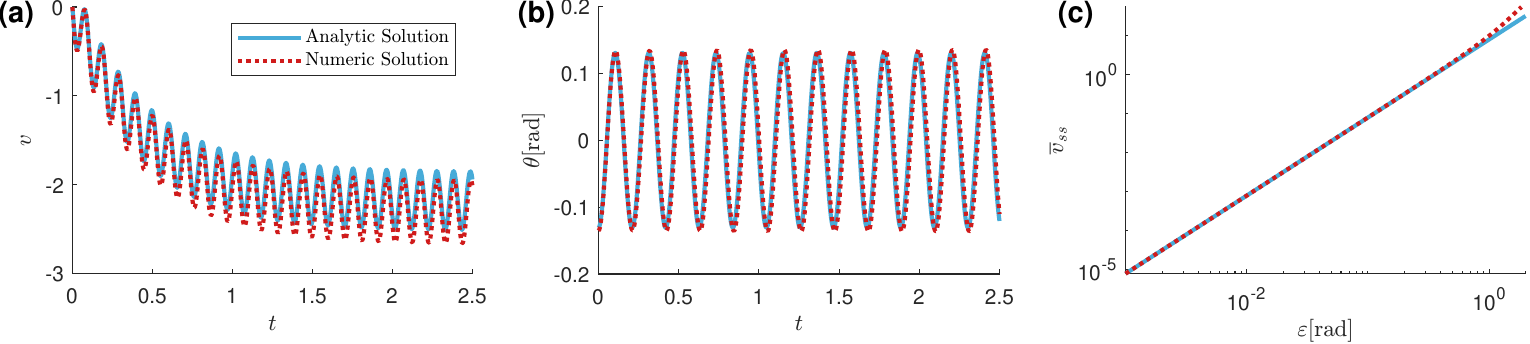}
\caption{Comparison between numeric simulation and the leading-order analytic approximation, using the parameters in \eqref{Input Angle}, \eqref{Simulation Parameters}. (a) Time plot of the vehicle's forward speed \(v(t)\). (b) Time plot of the orientation angle. (c) Log-log plot of the mean velocity in steady-state vs. the input amplitude.} \label{fig4}
\end{figure*}

In Figure~\ref{fig4}(a), we present the nondimensional analytic solution for the forward speed \(v(t)\) overlaid with the numeric solution, both obtained with the parameter values as in \eqref{Input Angle for simulations} and \eqref{Simulation Parameters}. We check the fit quality between the two solutions in Figure~\ref{fig4}(c), plotting the mean forward speed in steady-state for different input amplitudes on a log-log scale. The analytic solution is a linear line with a slope of 2 due to its leading order \(\varepsilon^2\), and we see a good alignment between this line and the numeric solution for \(\varepsilon \ll 1\) and even for input amplitudes which are closer to 1.

Next, we examine the conditions for direction reversal. We note that the sign of the mean value of the solution \(v(t)\) in steady-state, \(b_1\), dictates the direction of movement of the vehicle. In \eqref{Direction Reversal Condition} we show \(\xi\), the condition for direction reversal under the simplifying assumption \(\beta_2 = 1/2\). \(\xi\) is positively proportional to \(b_1\), consisting of the terms that dictate its sign. When \(\xi > 0\), the mean velocity of the vehicle is positive, and when \(\xi < 0\), the mean velocity of the vehicle is negative.
\begin{equation} \label{Direction Reversal Condition}
    \xi = 4 \beta_1 \left(\alpha - \beta_1\right) - 4 \eta_1 - \kappa \left( 2 - \alpha \right) + 4 \alpha \eta_2 \kappa
\end{equation}

We compare our results to the non-dissipative model in \cite{halvani_nonholonomic_2022}, in which the direction reversal condition is determined by the mean acceleration of the vehicle. Under the same assumption of \(\beta_2 = 1/2\), we obtain the same condition for direction reversal.

Using the relation in \eqref{Body Velocities Reduction}, we can restore the orientation angle of the body. First, we retrieve the orientation's angular velocity:
\begin{equation} \label{Symmetric - Angular velocity Solution}
    \theta'(t) = \frac{\sin{(\phi (t))}}{\alpha + \cos{(\phi (t))}} v(t) - \frac{\alpha}{\alpha + \cos{(\phi (t))}} \phi'(t)
\end{equation}

Then, we substitute the velocity's leading-term solution and the steering angle into \eqref{Symmetric - Angular velocity Solution}, and integrate with respect to time. We nullify the integration constant so that the steering angle oscillates around zero, and obtain:
\begin{equation} \label{Symmetric - Theta Solution}
    \theta(t) = - \left( \frac{\alpha}{1 + \alpha} \cos{(t \omega)} \right) \varepsilon + \mathcal{O}\left( \varepsilon^3 \right)
\end{equation}

In Figure~\ref{fig4}(b), we present the nondimensional leading-order orientation angle of the vehicle overlaid with the numeric solution, showing the match between the two. { It should be noted that the $\mathcal{O}\left( \varepsilon^3 \right)$-residual terms in \eqref{Symmetric - Theta Solution} contain only zero-mean oscillations, due to symmetry of the model and the input $\phi(t)$ with respect to $\phi=0$. This changes when the input has nonzero mean $\phi_0 \ne 0$, as discussed next.}

\subsection{Asymmetric input}
We consider an input with small deviation \(\phi_0 \ll 1\) from the symmetric case. We write \eqref{Asymptotic Analysis - Equation Form} in a nondimensional form and expand \(v'(t)\) in \eqref{Asymptotic Analysis - Nondimensional Equation Form} { with respect to both $\varepsilon$ and $\phi_0$}. As a result, we obtain { leading order terms in $\varepsilon$ and $\phi_0$ of the solution for $v(t)$, as} 
\begin{equation} \label{Asymmetric - Velocity Expansion}
    v(t) = \varepsilon^2 v_2(t) + \varepsilon \phi_0 v_{10}(t) + \mathcal{O}\left( {\phi_0}^2 \varepsilon, \phi_0 \varepsilon^2, \varepsilon^3 \right)
\end{equation}
where \(v_2(t)\) is the second-order approximation, identical to the symmetric case \(\phi_0=0\) we showed in \eqref{Symmetric - Velocity Solution}, and \(v_{10}(t)\) is a correction term given by:
\begin{equation} \label{Asymmetric - Velocity Solution}
    v_{10}(t) = c_2 \left( \sin{\left(\omega t \right)} - \mathrm{e}^{-\frac{3t}{1 + \kappa}} \right) + c_3 \cos{\left(\omega t \right)}
\end{equation}
where the expressions for \(c_2, c_3\) appear in Appendix \hyperref[tab2]{A}.


\begin{figure*}[!t]
\centering
\includegraphics[width=\textwidth, clip]{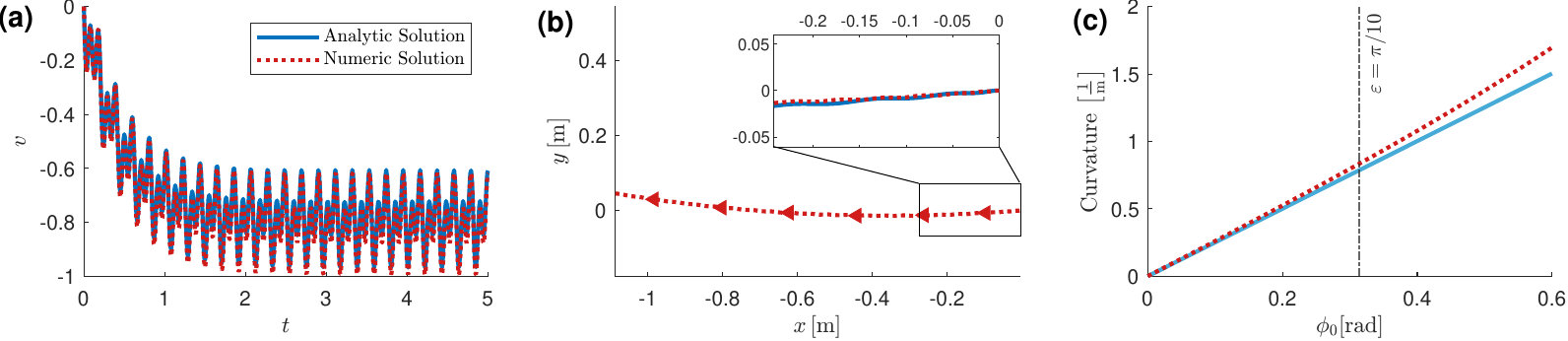}
\caption{Comparison between numeric simulation and analytic approximation using \eqref{Simulation Parameters} and \(\varepsilon = \pi / 10 \mathrm{[rad]}\) (a) Nondimensional forward speed comparison with asymmetry \(\phi_0 = 5^{\circ}\). { (b) The robot's trajectory in $(x,y)$ plane under $\phi_0 = 5^{\circ}$, resulting in curved arc. The inset shows a zoomed-in comparison with the leading-order analytic approximation of the path.} (c) Dimensional mean curvature of the robot's path,  { comparison between the numerical solution in dahsed red line and leading-order approximation in blue line.} The dashed-dotted vertical line is the value of the input angle's amplitude \(\varepsilon\).} \label{fig5}
\end{figure*}

The correction term in \eqref{Asymmetric - Velocity Solution} oscillates around zero at steady-state, meaning it does not affect the direction reversal condition. In addition, it oscillates at the input frequency, in contrast to \(v_2(t)\) which oscillates at twice this frequency. In Figure~\ref{fig5}(a), we show a comparison between the numeric simulation and analytic approximation of the forward speed \(v(t)\) using the parameters in \eqref{Simulation Parameters}, while incorporating an asymmetric mean input angle of \(\phi_0 = \pi/36\mathrm{[rad]}\).

\begin{table*}[!b]
\caption{Physical values for different robot's configurations}\label{tab1}
\begin{tabular*}{\textwidth}{@{\extracolsep\fill}lcccccccc}

 & \(m_1\) & \(m_2\) & \(l_1\) & \(l_2\) & \(b_1\) & \(b_2\) & \(J_1\) & \(J_2\)\\
 & \(\mathrm{[Kg]}\) & \(\mathrm{[Kg]}\) & \(\mathrm{[m]}\) & \(\mathrm{[m]}\) & \(\mathrm{[m]}\) & \(\mathrm{[m]}\) & \(\mathrm{[Kg \cdot m^2]}\) & \(\mathrm{[Kg \cdot m^2]}\) \\ 

\hline

Nominal Robot & 0.836 & 0.29 & 0.144 & 0.112 & 0.0206 & 0.068 & 0.0636 & 0.003873\\
Backward Configuration  & 0.836 & 0.383 & 0.144 & 0.208 & 0.0206 & 0.098 & 0.0636 & 0.01938\\
Forward Configuration  & 1.8 & 0.383 & 0.144 & 0.208 & 0.0426 & 0.098 & 0.07757 & 0.01938\\
\hline
\end{tabular*}
\end{table*}

We use \eqref{Symmetric - Angular velocity Solution} to find the orientation angle's velocity \(\theta'(t)\) by substituting the velocity as in \eqref{Asymmetric - Velocity Expansion} and expanding as a power series in \(\varepsilon\). We show the steady-state solution of the angular velocity in \eqref{Asymmetric - Angular Velocity Solution (ss)}. This solution contains a mean value that appears due to the asymmetric input. This constant change in the mean orientation angle of the vehicle leads to a motion along a mean circular arc.
\begin{multline} \label{Asymmetric - Angular Velocity Solution (ss)}
    \theta'_{ss}(t) = \varepsilon d_0 \sin{(\omega t)} \\ + \varepsilon^2 \phi_0 \left(d_1 + d_2 \sin{(2\omega t)} + d_3 \cos{(2\omega t)}\right) + \mathcal{O}\left( {\phi_0^2} \varepsilon\right)
\end{multline}
where expressions for \(d_0,d_1,d_2,d_3\) appear in Appendix \hyperref[tab2]{A}.

{ Equation \eqref{Asymmetric - Angular Velocity Solution (ss)} implies that the robot's path now oscillates around  a circular arc.} We derive the approximate curvature of the robot's mean path using the asymmetric analytic solutions in \eqref{Asymmetric - Velocity Expansion} and \eqref{Asymmetric - Angular Velocity Solution (ss)}:
\begin{equation} \label{Approximated Curvature}
    \rho = \frac{\overline{\theta_{ss}'}}{\overline{v}} = \frac{\varepsilon^2 \phi_0 d_1 + \mathcal{O}\left({\phi_0}^2 \varepsilon\right)}{\varepsilon^2 b_1 + \mathcal{O}\left( {\phi_0}^2 \varepsilon, \phi_0 \varepsilon^2, \varepsilon^3 \right)} \approx \frac{d_1}{b_1} \phi_0
\end{equation}

{ Figure ~\ref{fig5}(b) plots the robot's trajectory in $(x,y)$ plane under \(\varepsilon = \pi / 10 \mathrm{[rad]}\) and  \(\phi_0 = 5^{\circ}\), resulting in curved arc. The inset shows a zoomed-in comparison with the leading-order analytic approximation of the path, in blue curve.}
In Figure~\ref{fig5}(c), we present a comparison between the numeric and approximate curvatures { in a log-log plot as a function of $\phi_0$. This comparison  shows that for relatively small input asymmetry  \(\phi_0 < \varepsilon\)}, left of the dashed-dotted vertical line, the curvature is linearly proportional to the mean angle value. { on the other hand, for \(\phi_0 > \varepsilon\), residual terms in \eqref{Approximated Curvature} are not small,} and the discrepancy becomes significant. 

\section{Experiments} \label{sec5}
\begin{figure}[!h]
\centering
\includegraphics[width=0.5\textwidth, clip]{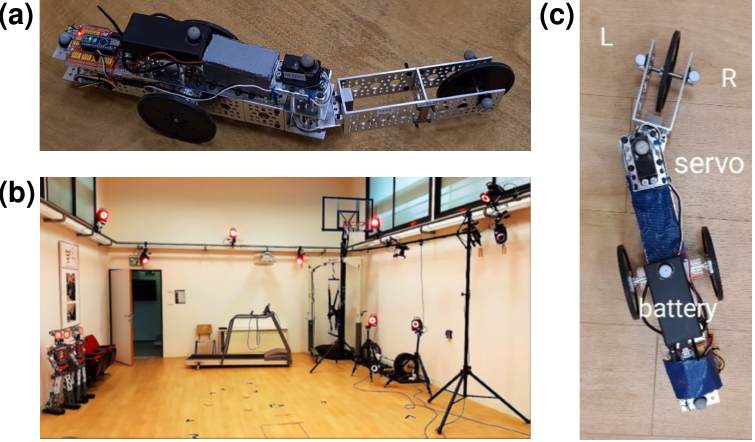}
\caption{The robot and experimental setup. (a) The robot in the `Forward Configuration', with the mass block positioned on link 1. (b) VICON camera setup, camera locations are marked with red dots. (c) The four tracking markers, positioned on the robot.} \label{fig6}
\end{figure}

For the experiment, we designed a robot with a modular reconfigurable geometric structure and mass distribution, as shown in Figure~\ref{fig6}(a). We used a goBILDA kit for the chassis, a servo motor for the relative angle actuation, and an Arduino microcontroller for controlling the servo's movement.

The experiments were conducted using three main robot configurations, with physical parameter values as shown in Table~\ref{tab1}. Those values were taken from the robot's CAD model. The `Nominal Robot' configuration is the first and basic version of the robot, which was used for the parameter fitting experiments. Then, we modified the robot's geometry so that a direction reversal would be possible by adding a block mass \(m_0\) onto link 1. In the `Backward Configuration', we extended the length of link 2 so that the forward speed of the vehicle is still negative but close to zero. The `Forward Configuration' is the same configuration while adding the mass block, incorporating the block mass and moment of inertia into the properties of link 1.

\subsection{The experimental setup}
We performed motion tracking using the VICON measurement system, using a sampling rate of 120 Hz. This system is equipped with several cameras capable of tracking the location of reflective markers. Both camera setup and the position of the marker on the robot are presented in Figures~\ref{fig6}(b)-(c). We use Python code to filter the results and calculate the coordinates and velocity of the robot.

\subsection{Parameter fitting} \label{sec5B}
To determine the value of the unknown viscous dissipation coefficient, we conduct a parameter fitting. In this section, all experiments were conducted using the `Nominal Robot' configuration, as presented in Table~\ref{tab1}, under open-loop input.

\begin{figure*}[!t]
\centering
\includegraphics[width=\textwidth,clip]{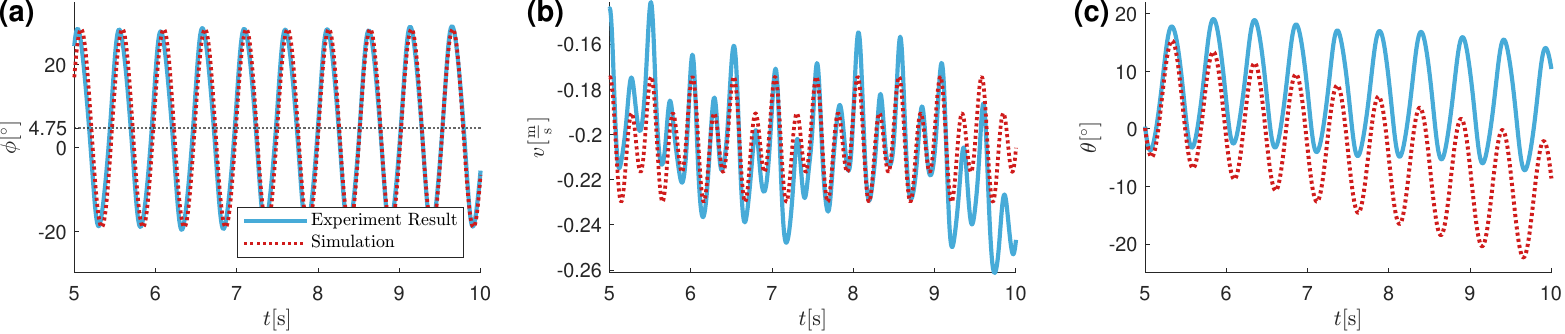}
\caption{Comparison between experimental results and simulation results with fitted viscous dissipation coefficient of \(c = 0.234 \mathrm{[N \cdot s / m]}\). (a) Relative angle between the two links. (b) Vehicle's body-frame velocity. (c) Orientation angle.} \label{fig7}
\end{figure*}

First, we look at a specific experiment with an input frequency of 14 radians per second and amplitude of \(\pi / 6\) radians, shown in Figure~\ref{fig7}. From Figure~\ref{fig7}(a), we understand that the tracked relative angle \(\phi(t)\) is different from the commanded input angle by its mean value, amplitude, and frequency. { Possible reason for this discrepancy is imperfections of the chosen servo motor in our robot and its internal dynamic response. Deviation in the mean value of the input angle might be caused by unmodelled effects of mechanical backlashes, friction and misalignment of the joint’s axis with vertical direction.

In order to compensate for this discrepancy,} we define the tracked relative angle parameters in \eqref{Experiments - Input Angle Form}.
\begin{equation} \label{Experiments - Input Angle Form}
    \Phi(t) = \Phi_{\text{Mean}} + \Phi_{\text{Amp}} \cos{\left( \Omega t \right)}
\end{equation}

In this experiment, we find the tracked input angle's parameters: \(\Phi_{\text{Mean}} = 4.75 ^{\circ}\), \(\Phi_{\text{Amp}} = 23.75 ^{\circ}\) and \(\Omega = 12.36 \mathrm{[rad / s]}\), obtained by using the results' extremum points. We conduct a numerical simulation under the tracked input while hand-tuning the viscous dissipation coefficient. We add to Figure~\ref{fig7} the simulation results with the best-fitted coefficient of \(c = 0.234 \mathrm{[N \cdot s / m]}\). Using this model, we were able to capture the behavior of the robot's forward speed \(v(t)\) to a great extent, but were not able to match the actual drift rate of the orientation angle \(\theta(t)\). { This is in spite of incorporating asymmetric input of steering angle with nonzero mean $\phi_0$. A possible explanation for this discrepancy is some structural asymmetry in the wheels’ lateral friction due to out-of-plane effects such as deviation of the wheel’s planes from vertical, as well as suspension unit assembled at the double-wheel rear axle. This structural asymmetry seems to counteract the asymmetry induced by the input.}

We note that in Figure~\ref{fig7} we take a specific time window from the experimental measurements. Outside of this time window, the vehicle does not reach a steady-state motion and behaves irregularly. This behavior can be explained by external effects that the robot is sensitive to, and we did not account for while performing the experiment. In the following analysis, we look at each experiment and try to identify a steady-state, such as in Figure~\ref{fig7}(b), which we can analyze and compare to simulations.

Next, after determining a suitable viscous dissipation coefficient \(c\) for a specific experiment, we evaluate its applicability across a variety of experiments with varying input frequencies. We performed approximately 20 experiments with an input amplitude of 30 degrees and varying input frequencies between 6 and 15 rad/s. 

\begin{figure}[!h]
\centering
\includegraphics[width=.47\textwidth,clip]{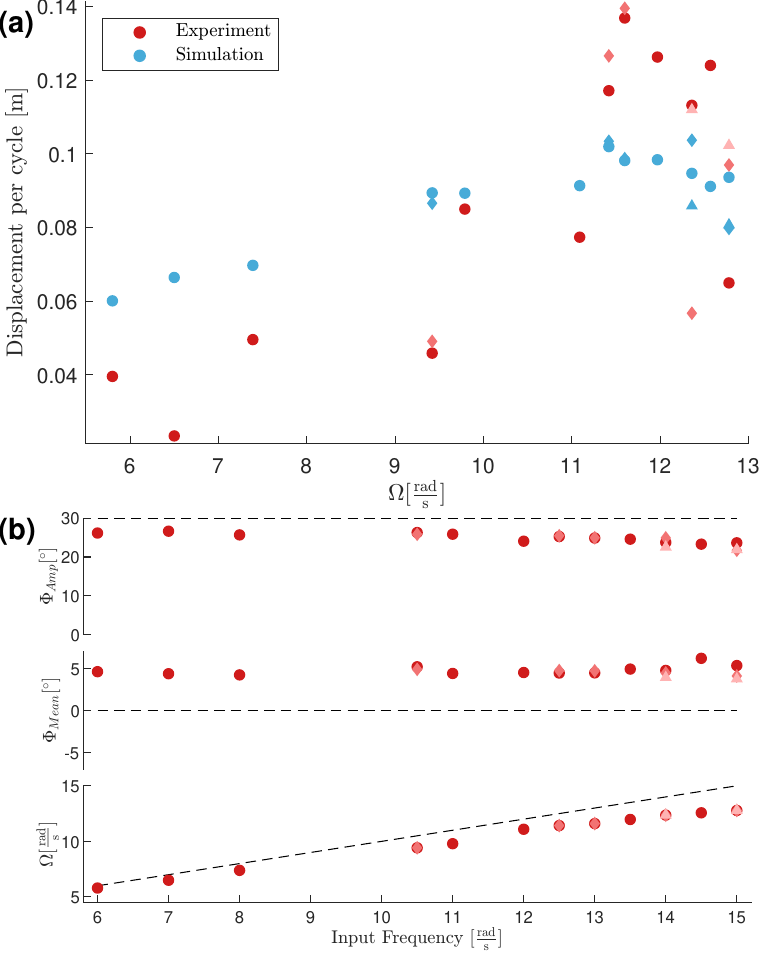}
\caption{Dissipation coefficients estimation using frequency-dependent results. (a) Displacement per cycle of different input frequency experiments overlaid with simulation results using \(c = 0.4 \mathrm{[N \cdot s / m]}\). (b) Captured steering angle properties for different experiments, in dashed line we plot the expected steering angle properties.} \label{fig8}
\end{figure}

For each experiment, we measure the actual steering angle \(\phi(t)\), shown in Figure~\ref{fig8}(b), which deviates significantly from the commanded input values. We use the actual steering angles as the simulations' input, and use the results to compare with the experiments. In Figure~\ref{fig8}(a), we show the mean displacement per cycle of each experiment. By hand-tuning the viscous dissipation coefficient, we achieve the best alignment between the experiments and simulations for \(c=0.4 \mathrm{[N \cdot s / m]}\).

From the comparison, we notice that no damping coefficient value matches the results at both high and low input frequencies. In addition, experiments conducted at high frequencies exhibit significant variance between different experiments under the same input. These results highlight the sensitivity of the robot and complicate the process of reliable parameter fitting.

\subsection{Direction reversal experiment} \label{sec5.4}
\begin{figure}[!h]
\centering
\includegraphics[width=.48\textwidth,clip]{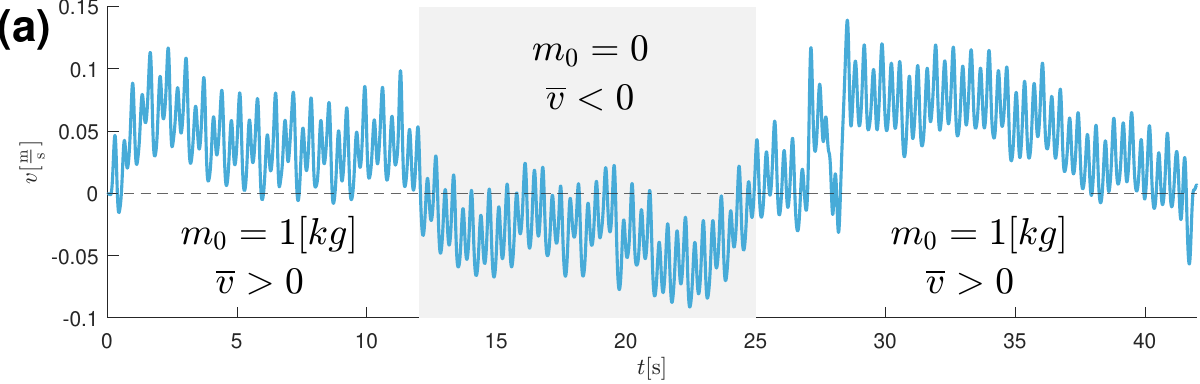}\\
\includegraphics[width=.48\textwidth,clip]{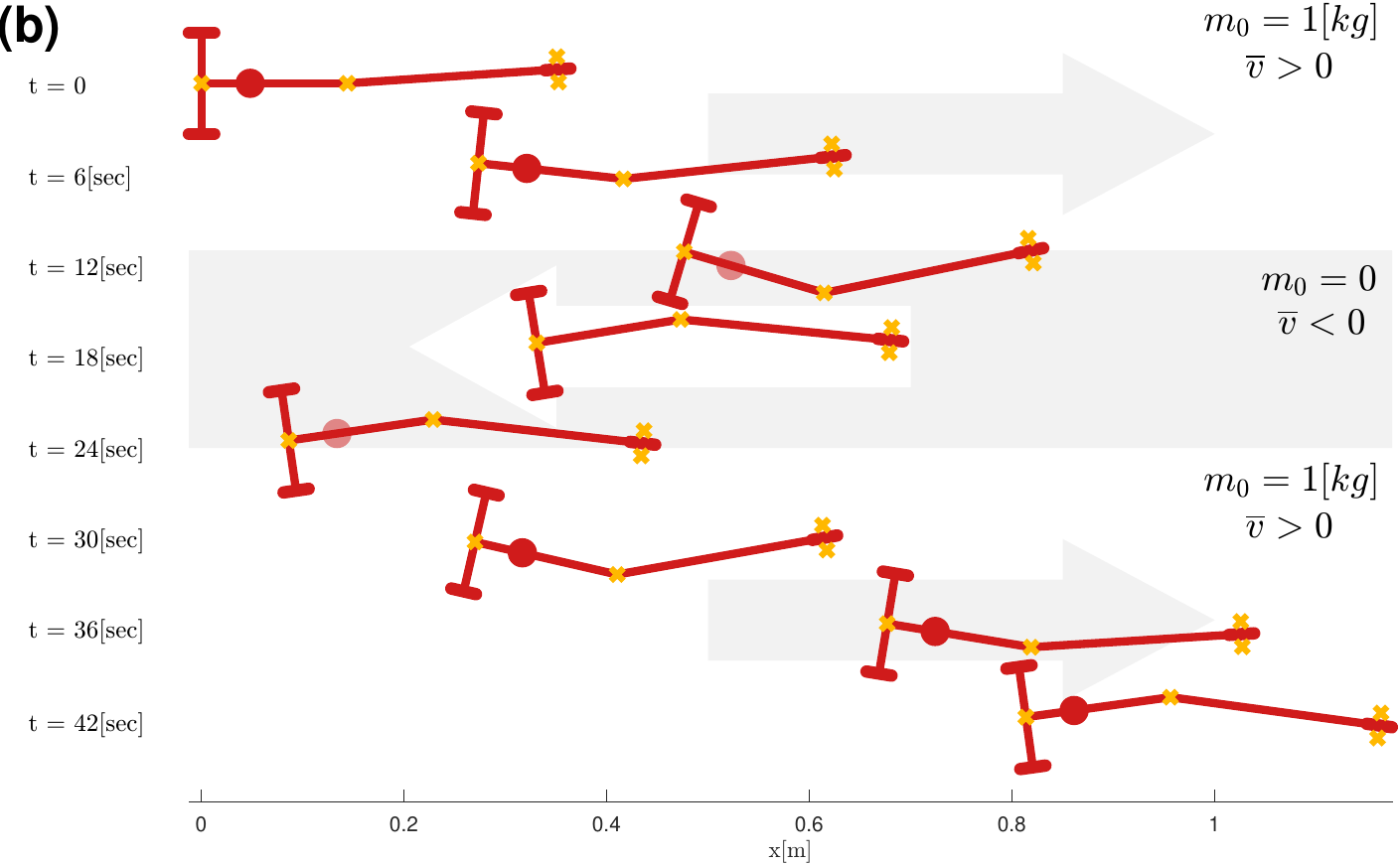}
\caption{Direction reversal experiment results, white background - the robot is in `Forward Configuration', gray background - the robot is in `Backward Configuration'. (a) Robot speed \(v(t)\) measurements. (b) snapshots appearing at fixed time intervals of 6 seconds.} \label{fig9}
\end{figure}

Next, we obtain a direction reversal of the robot's movement by placing a mass block on it. In Figure~\ref{fig9}, we present the experiment results as captured using the tracking system. The results are also shown in a supplementary movie\footnote{This paper has supplementary downloadable material available at \url{https://drive.google.com/file/d/1BL6pWKHyYNvPq4p_D62BVZR0ibrW55cn/view?usp=sharing}, provided by the authors. This includes a theoretical and experimental overview of the direction reversal phenomenon. This material is 67.6 MB in size.}. Figure~\ref{fig9}(a) contains the robot's forward speed versus time, while Figure~\ref{fig9}(b) contains snapshots of the robot appearing at fixed time intervals of 6 seconds.

\begin{figure*}[!b]
\centering
\includegraphics[width=\textwidth,clip]{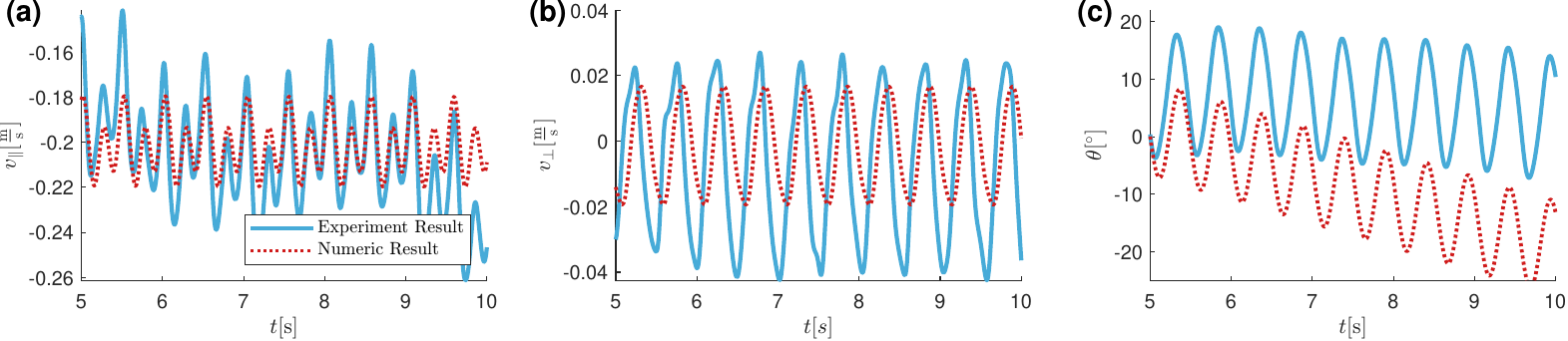}
\caption{Comparison between experimental results and simulation results with fitted viscous dissipation coefficients of \(c_{\parallel} = 0.095[N \cdot s / m]\), \(c_{\perp} = 4 [N \cdot s / m]\). (a) Vehicle's body-frame rolling velocity. (b) Vehicle's body-frame skid velocity. (c) Orientation angle.} \label{fig10}
\end{figure*}

The robot starts in the `Forward Configuration', including the mass block. In the gray sections, we show the time intervals in which the robot is in the `Backward Configuration', when we take the block off.

From the results, we see that direction reversal is possible by changing the mass distribution of the robot. Note that the human interaction with the robot at the beginning, the ending, and the transition phases between configurations has some effect on the results. Nevertheless, we see a clear difference between forward and backward movement while changing the configuration.

\section{Discussion of additional effects} \label{sec6}
From the comparison results in Figure~\ref{fig7}(c) and Figure~\ref{fig8}(a), we see that the model does not capture the behavior of the experimental robot in full. We try to explain the differences by demonstrating the effect of two phenomena on the robot's behavior. The first is the wheels' skidding, and the second is the effect of a slope on the model, which we investigate its impact on the model's behavior.

\subsection{Wheels' skidding}
In Figure~\ref{fig10}(a,c), we show the results of the same experiment as in Figure~\ref{fig7}, and add in Figure~\ref{fig7}(b) the measured skid velocity \(v_{\perp}(t)\) of the rear wheels. The skid velocity of the vehicle is smaller by one order of magnitude than the forward speed, meaning it is small but not completely negligible.

In the following analysis, we incorporate wheels' skid into the model by replacing the nonholonomic no-skid constraints with skid dissipation. { For simplicity and convenience, we assume viscous skidding friction, in order to avoid non-smoothness and numerical sensitivities incurred by using Coulomb's friction \cite{yona_wheeled_2019,halvani_nonholonomic_2022}.} We use the Jacobian forms in \eqref{Skid Velocity - with Jacobian} and \eqref{Roll Velocity - with Jacobian}, and write the Rayleigh's dissipation function when \(c\) is the viscous dissipation coefficient at the roll direction of each wheel and \(c_{\perp}\) is the viscous dissipation coefficient at the skid direction:
\begin{multline} \label{Skiding Discussion - Rayleigh Dissipation Function}
    \resizebox{!}{.95em}{\( \mathcal{R}(\textbf{q}, \dot{\textbf{q}}) = \frac{c}{2} \sum_{i=1}^{3}{\left( {\textbf{v}_{\parallel, i}} \right)^{T} \left(\textbf{v}_{\parallel, i} \right)} + \frac{c_{\perp}}{2} \sum_{i=1}^{3}{\left( {\textbf{v}_{\perp, i}} \right)^{T} \left(\textbf{v}_{\perp, i} \right)} \)} \\
    \resizebox{!}{.95em}{\(= \frac{1}{2} {\dot{\textbf{q}}}^T \left( c \sum_{i=1}^{3}{{\mathbb{J}_i(\textbf{q})}^T \mathbb{J}_i(\textbf{q})} + c_{\perp} \sum_{i=1}^{3}{{\mathbb{J}_{\perp, i}(\textbf{q})}^T \mathbb{J}_{\perp, i}(\textbf{q})} \right) \dot{\textbf{q}} \)}
\end{multline}

We use the Euler–Lagrange method as in \eqref{Euler-Lagrange Constrained Equation}, while nullifying the constraints.
\begin{equation} \label{Skidding Discussion - Dynamic Equations - Matrix Form}
    \textbf{M}(\textbf{q}) \ddot{\textbf{q}} + \textbf{B}(\textbf{q}, \dot{\textbf{q}}) + \textbf{D}(\textbf{q}, \dot{\textbf{q}}) = \textbf{F}_q
\end{equation}
where \(\textbf{M}(\textbf{q})\) and \(\textbf{B}(\textbf{q}, \dot{\textbf{q}})\) are derived the same way as in \eqref{Dynamic Equations - Matrix Form} and:
\begin{equation} \label{Dissipation Vector - Discussion}
    \textbf{D}(\textbf{q}, \dot{\textbf{q}}) = \frac{\partial \mathcal{R}}{\partial \dot{\textbf{q}}}
\end{equation}

To conduct the simulation, we divide the matrices into passive and active coordinates the same way as in Section~\ref{sec3}, and rearrange into matrix form when the unknown vector consists of the passive coordinates' acceleration and the input torque.
\begin{equation} \label{Skidding Discussion - Dynamic Equations - Simulations}
    \begin{bmatrix}
        \textbf{M}_{pp} & \textbf{0}^{3 \times 1} \\
        {\textbf{M}_{pa}}^T & -1
    \end{bmatrix} \begin{bmatrix}
        \ddot{\textbf{q}}_p \\ \tau
    \end{bmatrix} = - \begin{bmatrix}
        \textbf{M}_{pa} \ddot{\textbf{q}}_a + \textbf{B}_p + \textbf{D}_p \\
        \textbf{M}_{aa} \ddot{\textbf{q}}_a + \textbf{B}_a + \textbf{D}_a
    \end{bmatrix}
\end{equation}

We hand-tune the viscous dissipation coefficients, and choose the best-fitted coefficients \(c_{\parallel}=0.095\mathrm{[N \cdot s / m]}\) and \(c_{\perp}=4 \mathrm{[N \cdot s / m]}\), as presented in Figure~\ref{fig10}. The rolling dissipation coefficient obtained in this section is much smaller than the one that was obtained using the constrained model. This difference may explain the difference between the simulation and the experiments.

\subsection{Uphill Slope}
While experimenting, we noticed that experiments with the same input frequencies behave differently, as shown in Figure~\ref{fig8}(a). We check if the inconsistencies of the levelness of the ground's surface on which the experiment was conducted have a significant influence on the results.

\begin{figure}[!h]
\centering
\includegraphics[width=0.49\textwidth,clip]{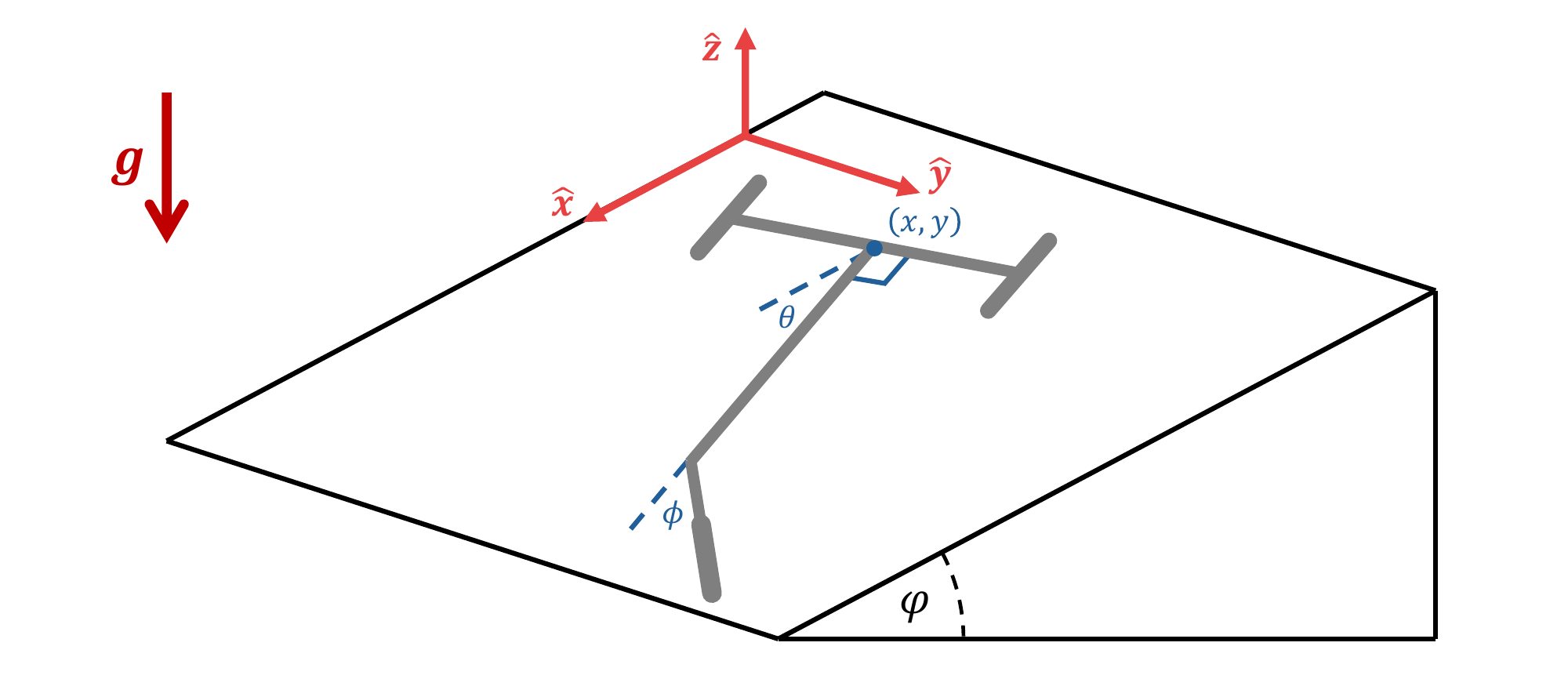}
\caption{Gravitational-planar model of the Twistcar.} \label{fig11}
\end{figure}

We model this as an uphill slope of angle \(\varphi\), as we show in Figure~\ref{fig11}. We take the same constrained-dissipative planar model we showed in Figure~\ref{fig1}(a), and incorporate potential energy into it. For this section, we take \(m_0 = 0\) and the Potential energy is defined as:
\begin{equation*} \label{eq30}
    V = \textbf{g} \cdot \left( m_1 \textbf{r}_{c_1} + m_2 \textbf{r}_{c_2} \right), \quad \textbf{g} = g \begin{bmatrix} -\sin{\varphi} && 0 \end{bmatrix}
\end{equation*}
We add to \eqref{Dynamic Equations - Matrix Form} the gravitational term \textbf{G}:
\begin{align}
    \textbf{M}(\textbf{q}) \ddot{\textbf{q}} + \textbf{B}(\textbf{q}, \dot{\textbf{q}}) + \textbf{D}(\textbf{q}, \dot{\textbf{q}}) + \textbf{G}(\textbf{q}) &= \textbf{F}_q + \textbf{W}^T(\textbf{q}) \boldsymbol{\Lambda} \nonumber \\
    \textbf{W}(\textbf{q}) \ddot{\textbf{q}} + \dot{\textbf{W}}(\textbf{q}) \dot{\textbf{q}} &= 0 \label{Uphill Slope Discussion - Dynamic Equations - Matrix Form}
\end{align}
where \(\textbf{M}(\textbf{q})\), \(\textbf{B}(\textbf{q}, \dot{\textbf{q}})\) and \(\textbf{D}(\textbf{q}, \dot{\textbf{q}})\) are identical to the matrices used in \eqref{Dynamic Equations - Matrix Form}, and:
\begin{equation*}
    \textbf{G} = \frac{\partial V}{\partial \textbf{q}} = \resizebox{!}{2.3em}{ \(\begin{bmatrix} -g (m_1 + m_2)\sin{\varphi} \\ 0 \\ g\sin{\varphi} [(b_1 m_1 + l_1 m_2) \sin{\theta} + b_2 m_2 \sin{(\theta + \phi)}] \\ g b_2 m_2 \sin{(\theta + \phi)} \sin{\varphi} \end{bmatrix} \)}
\end{equation*}

We simulate the gravitational model with the `Nominal Robot' configuration from Table~\ref{tab1} { with viscous rolling resistance coefficient of \(c=0.5 \mathrm{[N \cdot s / m]}\), under} a symmetric input with \(\Phi_{\text{Amp}} = 23.75^{\circ}\) and \(\Omega = 12.36 \mathrm{[rad / s]}\). We conduct the simulation three times. The first simulation is with a slope angle of \(\varphi=0\), which is equivalent to the model we used so far. In comparison, we conduct two more simulations with slope angles of \(\varphi=0.5^{\circ}\), shown in red, and \(\varphi=1^{\circ}\), shown in yellow. 

\begin{figure}[h]
\centering
\includegraphics[width=.49\textwidth,clip]{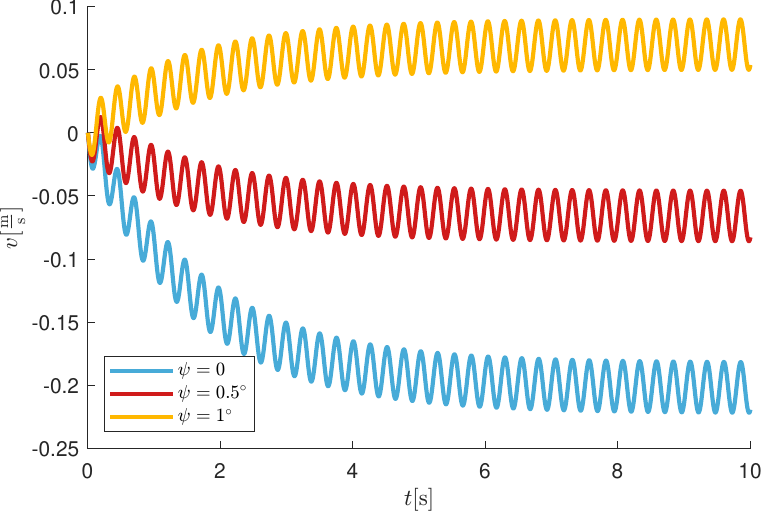}
\caption{Comparison between uphill slope simulations with different slopes, for the `Nominal Robot' configuration.} \label{fig12}
\end{figure}

In Figure~\ref{fig12} we demonstrate how significantly the levelness of the ground's surface influences the robot's speed, capable of changing the movement direction of the vehicle for a slope angle as small as \(1^{\circ}\). This result may explain the significant variance observed in Figure~\ref{fig8} between experiments conducted under the same input, as the ground conditions were not a primary focus during the experiments.

\section{Conclusions} \label{sec7}
In this work, we used theoretical analysis in order to investigate the dynamic behavior of the dissipative Twistcar and compared it to experiments. Both theoretical work and experiments showed that in the dissipative Twistcar, a direction reversal is obtainable by changing its geometry and mass distribution.

The analysis involved perturbation expansion and was validated using numeric simulations. Based on the theoretical analysis, we designed a modular robot demonstrating direction reversal. To determine the unknown viscous dissipation coefficient of the model, we conducted multiple experiments over a wide range of input frequencies, and also ran our simulation with input that was modified to match the actual measurements of joint angle in the experiment. By hand-tuning this coefficient, we found the best-fitting one as measured experimentally.

At the end of the work, we discussed some limitations of the model, as well as the influence of some effects that have not been accounted in the model. First, we incorporated the effect of wheels' skidding into the numerical simulations, which improved agreement with experimental measurements. Second, we showed that the levelness of the ground's surface can greatly influence the behavior of the robot.

Some future work includes improving the experimental setup. By utilizing a better controller { for the joint's servo motor}, it will be possible to obtain a more reliable input angle with a symmetric input. Such a controller will be able to input a wider range of frequencies, possibly capturing a direction reversal depending on the input's amplitude, as was shown in \cite{halvani_nonholonomic_2022, bazzi_motion_2017}.

Another future approach may involve the dissipation model that was used. The viscous dissipation model is a simple model that does not depend on normal ground reaction forces, for example. Incorporating more generalized friction models, { Such as Coulomb's friction and its smooth approximations} may lead to a better alignment between the experimental and theoretical work.

We also suggest using our approach to analyze other models using perturbation expansion and comparing it to experiments. For example, this can be done to the three-linked snake with one passive joint \cite{raz2026tro}, the rotor-actuated passive-steering Twistcar \cite{zigelman_dynamics_2025}, and other vehicles with passive joints \cite{rodwell_induced_2022}. 
Finally, one may consider applying methods of gait planning  { and optimization for achieving desired net motion of arbitrary maneuvers as in \cite{yang_geometric_2025}, rather than zero-mean sinusoidal inputs for generating straight-line net motion. In addition, the work can be extended to consider} motion planning and feedback stabilization \cite{grushkovskaya2023motion} to this underactuated nonholonomic system and its variants, { rather than focusing on open-loop dynamics}.

\begin{table*}[!b]
\captionsetup{name=APPENDIX A}
\renewcommand{\thetable}{}
\caption{explicit coefficients of the approximate analytic solution}
\label{tab2}

\begin{tabular*}{\textwidth}{@{\extracolsep\fill}l}

\hline

\(\Delta_1 = \left[6(\alpha + 1)^2\right]^{-1}\) \\ 
\(\Delta_2 = \left(4 (\kappa+1)^2 \omega ^2+9\right)^{-1}\) \\
\(\Delta_3 = \left((\kappa+1)^2 \omega ^2+9\right)^{-1}\) \\
\(\Delta_4 = [6(\alpha + 1)^3]^{-1}\) \\
\\
\(a_1 = -\alpha \omega^2 \left[ \alpha \beta_1 - {\beta_1}^2 + \kappa \left( \alpha + \beta_2 - 1 \right) - \eta_1 + \alpha \kappa \left( \eta_2 + \beta_2 \left(\beta_2 - 2\right) \right) \right] \Delta_1 a_2 \) \\
\(a_2 = - 3 / \left( \kappa + 1\right)\) \\
\(a_3 = \alpha \omega \left( \sigma^2 + 2 \alpha + 2 \right) \Delta_1 a_2 / 2 \) \\
\(a_4 = \alpha \omega^2 \left[ \alpha \beta_1 + {\beta_1}^2 + \kappa \left( \alpha + \beta_2 + 1 \right) + \eta_1 - \alpha \kappa \left( \eta_2 + \beta_2 \left(\beta_2 - 2\right) \right) \right] \Delta_1 a_2\) \\
\(b_2 = -2 \alpha  \omega^2 \left[2 \alpha  \beta _1 (\kappa +1)^2 \omega ^2+\alpha \beta _2^2 \kappa  \left(2 (\kappa +1)^2 \omega^2+9\right)+\left(2 (\kappa +1)^2 \omega^2+9\right) \left(\alpha  \eta _2 \kappa -\eta _1\right) + 2 \beta _2 \kappa  \left(-\left((2 \alpha -1) (\kappa +1)^2 \omega ^2\right)-9 \alpha \right) \right. \) \\
\hspace{20pt} \( + 3 \left(\alpha  \kappa +\alpha +2 (\kappa +1) \sigma ^2-2 \kappa +1\right) \left. +2(\alpha -1) \kappa  (\kappa +1)^2 \omega ^2-\beta _1^2 \left(2 (\kappa +1)^2 \omega ^2+9\right)\right] \Delta_1 \Delta_2\) \\
\( b_3 = 3 \alpha  \omega \left[2 (\kappa +1) \omega ^2 \left(\beta _2 \kappa  \left(\alpha  \beta _2-2 \alpha -1\right)-\alpha  \beta_1+\alpha  \eta _2 \kappa -\beta _1^2-\eta _1\right) -2 (\alpha +1) \kappa  (\kappa +1) \omega ^2  -3 \left(\alpha +2 \sigma^2+1\right)\right] \Delta_1 \Delta_2 \) \\
\( b_4 = 3 \alpha  \omega ^2 \Big[3 \kappa  \left(\beta _2 \left(\alpha  \beta _2-2 \alpha -1\right)+\alpha  \eta _2\right)-3 \alpha \beta _1-(\alpha +1) (\kappa -2) -3 \beta _1^2\Big. \)
\( \left. -3 \eta _1+4 (\kappa +1) \sigma ^2\right] \Delta_1 \Delta_2 \) \\
\\
\( c_2 = 6 \alpha  \omega ^2 \left(3 \alpha  \kappa  \left(\left(\beta _2-2\right) \beta _2+\eta _2\right)+\alpha  \kappa +\alpha -3 \beta _1^2-3 \eta _1+2 (\kappa +1) \sigma ^2-2 \kappa +1\right) \Delta_1 \Delta_3 \) \\
\( c_3 = 6 \alpha  \omega  \left((\kappa +1) \omega ^2 \left(\alpha  \kappa  \left(\left(\beta _2-2\right) \beta _2+\eta _2\right)-\beta _1^2-\eta_1\right)-3 \left(\alpha +2 \sigma ^2+1\right)-\kappa  (\kappa +1) \omega ^2\right) \Delta_1 \Delta_3 \) \\
\\
\( d_0 = \alpha \omega / (\alpha + 1) \) \\
\( d_1 = \alpha  \omega ^2 \Big[\alpha  \beta _2^2 \kappa  \left((\kappa +1)^2 \omega ^2+18\right)+\alpha  \beta _1 \left((\kappa +1)^2 \omega^2+9\right) +3 \left(4 \alpha  \kappa +\alpha +2 (\kappa +1) \sigma ^2-5 \kappa +1\right) +\beta _2 \kappa  \left(-\left((2 \alpha -1) (\kappa +1)^2 \omega ^2\right)-36 \alpha +9\right) \Big. \) \\
\hspace{20pt} \( +\left((\kappa +1)^2 \omega^2+18\right) \left(\alpha  \eta _2 \kappa -\eta _1\right) + (\alpha -1) \kappa  (\kappa +1)^2 \omega ^2-\beta _1^2 \left((\kappa +1)^2 \omega ^2+18\right)\Big] \Delta_3 \Delta_4 \) \\
\( d_2 = 3 \alpha  \omega \Big\{-(\kappa +1) \omega ^2 \left[2 \alpha  \beta _1 \left((\kappa +1)^2 \omega ^2+9\right)+2 \beta _2 \kappa  \left((6 \alpha +1) (\kappa +1)^2 \omega ^2+27 \alpha +9\right) -3 \alpha  \beta _2^2 \kappa  \left(2 (\kappa +1)^2 \omega ^2+9\right)\right. \Big. \)\\
\hspace{20pt} \( \left. +3 \left(2 (\kappa +1)^2 \omega ^2+9\right) \left(\eta _1-\alpha  \eta _2 \kappa \right)+3 \beta _1^2 \left(2 (\kappa+1)^2 \omega ^2+9\right)\right] +3 (\kappa +1) \omega ^2 \left(2 \alpha  (2 \kappa +5)-10 (\kappa +1) \sigma ^2+\kappa +10\right)\) \\
\hspace{20pt} \( +2 (\kappa+1)^3 \omega ^4 (\alpha  (\kappa +2)-\kappa +2) + 27 \left(\alpha -4 \sigma ^2+1\right)\Big\} \Delta_2 \Delta_3 \Delta_4\) \\
\(d_3 = 9 \alpha \omega ^2 \Big[ \alpha  \beta _2^2 \kappa  \left(5 (\kappa +1)^2 \omega ^2+18\right)-\alpha  \beta _1 \left((\kappa +1)^2 \omega ^2+9\right) -\beta _1^2 \left(5 (\kappa +1)^2 \omega^2+18\right) - \beta _2 \kappa  \left((10 \alpha +1) (\kappa +1)^2 \omega ^2+36 \alpha +9\right) \) \\
\hspace{20pt} \(+\left(5 (\kappa+1)^2 \omega ^2+18\right) \left(\alpha  \eta _2 \kappa -\eta _1\right) +(\kappa +1)^2 \omega ^2 \left(\alpha  (\kappa +2)+4 (\kappa +1) \sigma ^2-3 \kappa +2\right)+9 \left(\alpha +2 (\kappa +1) \sigma ^2-\kappa +1\right) \Big] \Delta_2 \Delta_3 \Delta_4 \) \\
\hline
\end{tabular*}
\end{table*}

\newpage

\section*{Acknowledgments}
The authors wish to express their gratitude to Professor Alon Wolf and the BRML team at the Technion for their support and for granting access to their laboratory facilities. In addition, sincere appreciation is extended to Ariel Bar-Yehuda, the lab engineer, for his invaluable help with the experiments. We also express our gratitude to Snir Carmeli and Dolev Freund for their efforts in conducting the experiments. { Finally, we wish to thank the anonymous reviewers for thorough reading and useful feedback.}

\clearpage

\bibliographystyle{IEEEtran}
\bibliography{bibliography}

\vfill

\end{document}